\documentclass{article}

\PassOptionsToPackage{numbers}{natbib}
\usepackage[preprint]{neurips_2024}
\usepackage{float}
\usepackage{graphicx}
\usepackage{balance}
\usepackage[table]{xcolor}
\usepackage{multirow}

\usepackage[utf8]{inputenc} 
\usepackage[T1]{fontenc}    
\usepackage{hyperref}       
\usepackage{url}            
\usepackage{booktabs}       
\usepackage{amsfonts}       
\usepackage{nicefrac}       
\usepackage{microtype}      
\usepackage{xcolor}         
\usepackage{subcaption}
\title{C3LLM: Conditional Multimodal
Content Generation Using Large Language Models}

\author{
    Zixuan Wang\textsuperscript{*}\\
    HKUST \\
    \texttt{zwanggk@connect.ust.hk} \\
    \And
    Qinkai Duan\textsuperscript{*}\\
    HKUST \\
    \texttt{qduanaa@connect.ust.hk}\\
    \AND 
    Yu-Wing Tai \\
    Dartmouth College \\
    \texttt{yuwing@gmail.com}
    \And
    Chi-Keung Tang \\
    HKUST \\
    \texttt{cktang@cse.ust.hk}\\   
}

\begin{document}

\maketitle
\newcommand{\ourwork}{C3LLM}

\begin{abstract}

We introduce C3LLM (Conditioned-on-Three-Modalities Large Language Models), a novel framework combining three tasks of video-to-audio, audio-to-text, and text-to-audio together. 
C3LLM adapts the Large Language Model (LLM) structure as a bridge for aligning different modalities, synthesizing the given conditional information, and making multimodal generation in a discrete manner. Our contributions are as follows. First, we adapt a hierarchical structure for audio generation tasks with pre-trained audio codebooks. Specifically, we train the LLM to generate audio semantic tokens from the given conditions, and further use a non-autoregressive transformer to generate different levels of acoustic tokens in layers to better enhance the fidelity of the generated audio. Second, based on the intuition that LLMs were originally designed for discrete tasks with the next-word prediction method, we use the discrete representation for audio generation and compress their semantic meanings into acoustic tokens, similar to adding ``acoustic vocabulary'' to LLM. Third, our method combines the previous tasks of audio understanding, video-to-audio generation, and text-to-audio generation together into one unified model, providing more versatility in an end-to-end fashion. Our C3LLM achieves improved results through various automated evaluation metrics, providing better semantic alignment compared to previous methods. 
\end{abstract}

\section{Introduction}
Conditional multimodal generation is the task of generating output that incorporates different modalities, such as text, image, video, and audio~\cite{Make-an-audio,CustomVideo,Imagen,Cogvideo}. Essentially, this multimodal task can be seen as a translation task among different modalities, and thus, challenges arise for making inferences from cross-modal representations and dealing with potential modality gaps~\cite{gap}. 

\begin{figure*}
  \centering
  \vspace{-0.1in}
  \includegraphics[width=\linewidth]{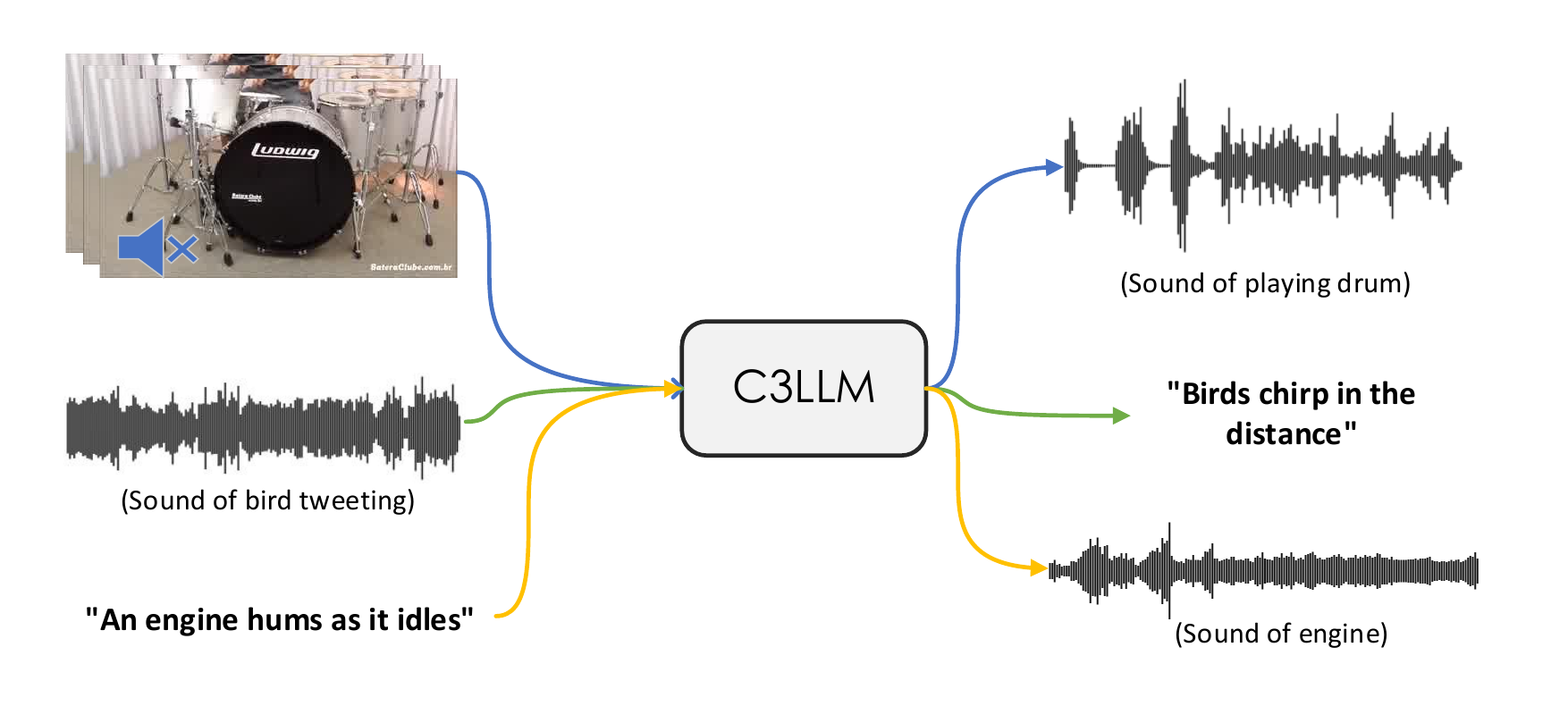}
  \vspace{-0.2in}
  \caption{C3LLM is capable  of video-to-audio, audio-to-text, and text-to-audio. Examples of different tasks are illustrated in arrows of different colors. 
  }
  \vspace{-0.2in}
  \label{fig:arch}
\end{figure*}

Multimodal Large Language Models (MM-LLMs) have recently gained significant interest in research due to their ability to understand and follow user instructions. Most work focuses on contextual understanding across various modalities like video-to-text~\cite{VLLAMA,AVLLM}, audio-to-text~\cite{LTU,audiogpt} and image-to-text~\cite{Visual-Instruction-Tuning}. However, the area of audio generation, particularly video-to-audio ~\cite{specvqgan,Conditional-floey} or image-to-audio generation~\cite{Im2Wav}, remains underexplored. This is partly because video contains excessive visual information not always needed for audio generation, while images lack the temporal information crucial for audio. Video-to-audio tasks also face synchronization challenges, with recent solutions like temporal masking~\cite{sonic} proving inadequate for complex scenarios. Additionally, current methods often encode video features by extracting a few random frames~\cite{sonic,clipsonic}, which hinders learning temporal information.


Audio generation is less intuitive compared to other tasks, as it is harder to precisely measure the generated sound quality using human ears. Additionally, previous works mainly focus on generating music-related audio, which is more structured compared to naturally occurring audio~\cite{musicgen,Mustango}. Few works have focused on generating visually guided open-domain audio clips~\cite{visually-aligned,Visual-to-Sound}. Moreover, most existing models~\cite{specvqgan,diffsound} are only limited to generating audio-only content, consequently constrained to specific downstream tasks. While contemporary work CoDi~\cite{codi} achieves some form of any-to-any generation, their result simply uses linear interpolations in the latent space. C3Net~\cite{c3net} adapted three ControlNet~\cite{controlnet} architectures on top of CoDi's design, still it also relies on interpolation when predicting the final output.  

In this connection, we aim at utilizing the versatility of LLM to align between different modalities. With sufficient data, transformers or LLMs have shown to be effective in serving as a unified backbone on different modality tasks even with simple design~\cite{emu}. We thus propose Conditioned-On-Three-Modalities Large Language Model, or~\textit{C3LLM} in short. C3LLM mainly comprises a LLM backbone serving as the bridge among three different modalities, and a hierarchy audio tokenizer adapted from Encodec~\cite{encodec} for decoding. Our model first encodes the respective modality, either audio, text or video, to be processed by the LLM backbone. For video, we extract the dense information and project it to the LLM's embedding space. For audio, we use the audio tokenizer to convert the information into discrete representation and translate the corresponding indices from the tokenizer codebooks into LLM special tokens, which are extended as part of LLM's vocabulary beforehand. The semantic information is further processed by the LLM. For tasks involving audio generation, We treat the LLM prediction as coarse acoustic tokens. The preliminary result is further extended to fine-grained acoustic tokens and combined to generate the final audio output, and thus 
multimodal output not limited to a single modality.

To sum up, our contributions are as follows:
1) C3LLM utilizes the versatility of LLM for conditional multimodal generation tasks, where the LLM treats encoded audio information as additional acoustic vocabulary; 
2) C3LLM uses a discrete tokenizer for modeling audio representation hierarchically, which better suits the nature of LLM while preserving the quality of the generated audio; 3) Consequently, this paper provides a uniform model for three different tasks involving video, audio and text, see Figure~\ref{fig:arch}. Through extensive evaluation, our work demonstrates on-par results compared to the state-of-the-art models in their respective domains, providing further insight into conditional multimodal tasks.

\begin{figure}
    \centering
    \includegraphics[width=\linewidth]{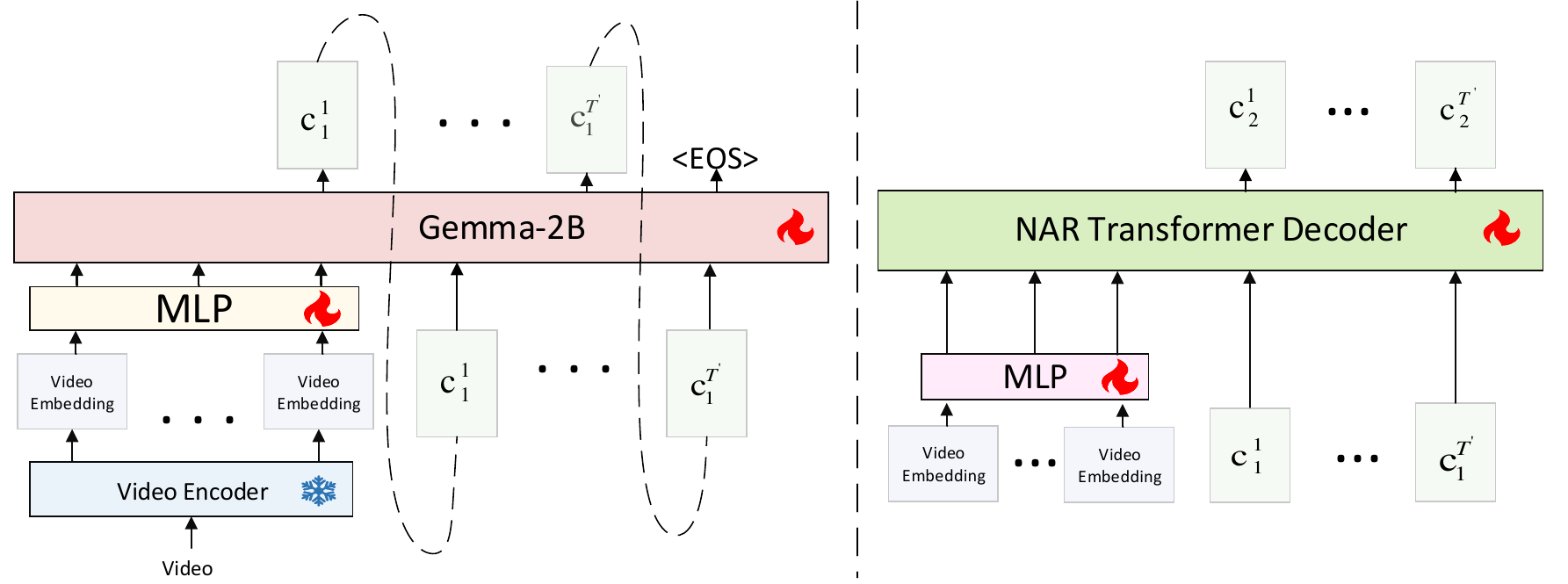}
    \vspace{-0.1in}
    \caption{Overview of the {\bf video} encoding part and the LLM. On the right is the non-autoregressive (NAR) transformer to further extend the coarse acoustic tokens into fine-grained acoustic tokens. 
    We freeze the encoder, train a MLP, and finetune LLM using LoRA.
    For text condition, we directly input them into LLM after tokenization and also treat them as the condition for NAR transformer decoding.
    }
    \label{fig2:env}
    \vspace{-0.1in}
\end{figure}
\section{Related Work}
\noindent\textbf{Multimodal Alignment}
Mokady et al.~\cite{clipcap} has utilized the powerful Contrastive Language-Image Pre-Training (CLIP)~\cite{clip} model to project the image into an image-text shared latent space. Given image embedding, they further apply a GPT-2~\cite{gpt2} to generate caption. 
Wu et al.~\cite{clap}  trained CLAP model on LAION-Audio-630K, a large collection of 633,526 audio-text 
pairs from different data sources, to obtain a robust and general result on all types of audio clips. 
Similar to them, we encode video, audio and text data into a shared latent space using contrastive learning and projection. This approach enables an audio encoder to project audio into a space shared with image and text, aligning more closely with the hidden space of a large language model. For long audio understanding, local features and overall information are fused.

\vspace{2mm}
\noindent\textbf{Multimodal LLM}
LLM is prevailing in the AI community. ChatGPT~\cite{chatgpt} and GPT-4~\cite{gpt4}
have shown great power in understanding and reasoning tasks. Other open source LLMs, such as LLama~\cite{llama}, Vicuna~\cite{vicuna}, Alpaca~\cite{alpaca}, and Gemma~\cite{gemma} 
have greatly contributed to the research community and inspired many innovations. With these advances, progress has been made in using LLMs for understanding multimodal information, improving the performance of MM-LLMs~\cite{Visual-Instruction-Tuning}.

We notice that most LLMs understand multimodal information with the help of a well-trained encoder to bridge the gap between modality information and LLM hidden space. For example, PandaGPT~\cite{pandagpt} utilizes ImageBind~\cite{imagebind} to align multiple modalities. Video-LLama~\cite{VLLAMA} applies two Q-formers to transform video and audio information. Llava~\cite{Visual-Instruction-Tuning} directly applies a linear projection layer to give LLM image information. MiniGPT-4 ~\cite{minigpt4} follows this approach, using linear layers to align Vicuna and Q-former. To fit the discrete and auto-regressive nature of LLMs, Large World Model~\cite{lwm} achieves long-context video understanding by combining VQGAN~\cite{vqgan} to tokenize each frame of the video. However, these LLMs lack the capability to generate modalities other than language.

\vspace{2mm}
\noindent\textbf{Multimodal Generation} In multimodal generation, the goal is to generate various modalities like audio, text, images, and videos interchangeably. State-of-the-art approaches, such as Composable Diffusion (CoDi)~\cite{codi} and C3Net~\cite{c3net}, produce diverse modality combinations conditioned on other modalities. NExT-GPT~\cite{nextgpt} and CoDi2~\cite{codi2} leverage LLMs to process and synthesize semantic information from different modalities, providing a wide range of meaningful conditions for the diffusion model. Diffusion models are crucial for generating and refining high-quality content, typically encoding each modality into a shared latent space and using MLPs to project information between the LLM hidden space and the diffusion latent condition space.

However, these methods often overlook the distinct features of each modality. For instance, NExT-GPT uses ImageBind~\cite{imagebind} to encode videos into image space, losing temporal features. In audio generation, the diffusion model struggles to map audio timing accurately to corresponding video frames, making it challenging to incorporate the temporal dimension of video modality.

\begin{figure}
    \centering
    \includegraphics[width=0.7\linewidth]{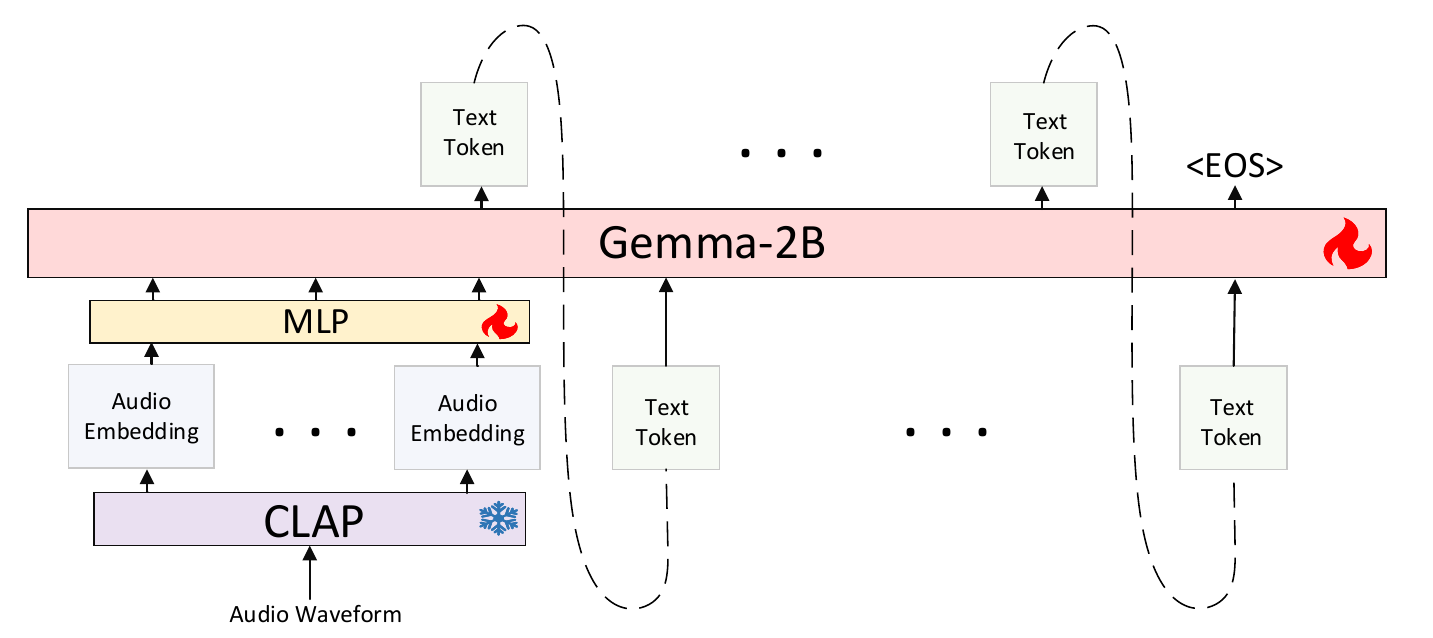}
    \vspace{-0.1in}
    \caption{Overview of {\bf audio} encoding part and the LLM. We use a pre-trained CLAP encoder and an MLP layer to align the audio information with LLM embedding.
    }
    \label{fig3:env}
    \vspace{-0.1in}
\end{figure}

\vspace{2mm}
\noindent\textbf{Conditioned Audio Generation} 
To better focus on the time domain of audio, the transformer architecture~\cite{transformer} is well-suited due to its attention mechanism. In the context of audio generation, SpecVQGAN~\cite{specvqgan} has successfully employed a neural codebook to represent audio information, enabling the use of a transformer to predict discrete audio tokens based on video features. 


Recently, VALL-E ~\cite{valle} and MusicLM~\cite{musiclm} use multiple codebooks and Residual Vector Quantization (RVQ)~\cite{encodec} to create diverse audio representations. An auto-regressive transformer is employed to predict the initial layers of audio indices based on the textual conditions, and a non-auto-regressive (NAR) transformer is followed to generate the subsequent layers, taking into account the previously generated tokens and conditions. It has been observed that the first audio index layer captures crucial information for overall audio quality, like rhythm and intonation, while subsequent layers add detailed information, enhancing richness. 


\section{Method}
C3LLM comprises three major components: 1) Tokenizers for encoding corresponding modality information to be comprehended by LLM; 2) a multimodal LLM that serves as a powerful backbone that connects corresponding modalities. Especially for audio generation tasks, we have 3) a non-autoregressive (NAR) Transformer that further refines the generated coarse acoustic tokens from the LLM. Our model structure is illustrated in Figure~\ref{fig2:env} and Figure ~\ref{fig3:env}.

\subsection{Audio Tokenizer and Video Tokenizer}
For audio generation, we employ the EnCodec~\cite{encodec} pretrained codebooks and decoder, which utilizes the RVQ method to obtain eight codebooks that enable high-bandwidth audio reconstruction. To streamline the process and conserve computational resources, we focus on the first two codebooks. We leverage the LLM to predict the first layer, taking into account text or visual conditions, and we train the NAR transformer decoder specifically to predict the second layer of audio tokens. After the prediction of the first two layers of audio tokens, the pre-trained decoder decodes them back into audio waveform.

For the audio-to-text task, we leverage the powerful audio encoder of CLAP to convert audio waveforms into a shared space that is more similar to the hidden space of the LLM. We freeze the audio encoder and introduce an MLP to project the audio vectors into the hidden space of the LLM. This allows for a more effective integration of audio information into the text-generation process. By utilizing the capabilities of CLAP and LLM together, we achieve improved performance in generating detailed descriptions from audio inputs.

When it comes to the video-to-audio task, we draw inspiration from the successful approach employed by SpecVQGAN ~\cite{specvqgan}. To efficiently capture both visual and temporal information while compressing the video data, we employ a frame-wise feature extractor denoted as $H$. This feature extractor extracts RGB features ($f^r$) and optical flow features ($f^o$) from each frame of the video. By applying frame-wise concatenation, we obtain the video feature representation $F = \{f_i^r,f_i^o\}_{i=1}^N$, where $N$ represents the total number of frames in the video. To further enhance the integration of visual information, we introduce an additional MLP that transforms the concatenated video features into embeddings suitable for the LLM.

Contrary to the discrete audio tokenization method, we choose to use the continuous representation for video. During our experiment, we observe that video involves too much visual information that audio generation may not necessarily need. A similar discrete method that uses VQGAN~\cite{vqgan} to process frame-level information will result in excessive visual tokens, making learning visual information inefficient. The MLP projection layer that we employed will project the continuous video feature into LLM embedding space, similar to previous methods~\cite{Visual-Instruction-Tuning}. For text input, we directly use the LLM's tokenizer for tokenization.

\subsection{Autoregressive Generation of Coarse Audio Tokens }\label{sssec:sec2.2}
Employing the EnCodec audio tokenizer allows us to represent continuous audio information in discrete form. We denote the continuous audio input as $a \in \mathbb{R}^{C \times L}$, where $C$ is the number of channels and $L$ is the time of the audio clip times sample rate. First, the audio input is encoded in a smaller representation in the form of $z = E(a) \in \mathbb{R}^{C \times N \times D \times Q}$, with $Q$ denoting the number of quantizers used during the encoding process and $D$ is the dimension of the codebooks. Our next step is to convert the representation into LLM-aware acoustic tokens. Specifically, we obtain the indices $s \in \mathbb{R}^{C \times N \times Q}$from the encoded audio by comparing with the quantizer codebook. 

To jointly model different modalities in a unified model, we further extended the LLM's text vocabulary $V_t = \{v_i\}_{i=1}^{N_t}$with acoustic vocabulary $V_a = \{v_j\}_{j=1}^{N_a}$. The extended audio-text vocabulary now becomes $V=\{V_t, V_a\}$. Contrary to previous audio generation models~\cite{specvqgan,Conditional-floey} that involve the generation of a single modality, our method equipped the LLM backbone with the ability to understand and generate both audio and text information with a unified vocabulary.

To better differentiate the three kinds of modalities that condition the autoregressive generation, we further wrap the encoded feature with special tokens as modality indicators. To be more specific, we wrap the audio tokens with \textit{<Audio>},\textit{</Audio>} indicators and video embedding in an embedded sequence of \textit{<Video>},\textit{</Video>} indicators. In doing so, we avoid the possibility of confusing the LLM with different kinds of information.   

To further elaborate on the conditional generation tasks performed by LLM: for audio-to-text and text-to-audio tasks, the source input $X_{a,t} = \{x_t^i\}_{i=1}^N$ is a sequence of either acoustic/text tokens. Here $N$ is the number of tokens we have and $x_t \in V$; for video-to-audio task, the source input $X_v = \{x_e^i\}_{i=1}^N$ is a sequence of embeddings and $x_e \in \mathbb{R}^D$, where $D$ is the embedding dimension of LLM. After LLM's tokenization, the input tokens for audio and text will become input embeddings and fed into the LLM. Our LLM backbone is a decoder-only structure with the next token prediction method. The distribution of the predicted token in the first layer is given by $p_{\theta_{LLM}}(\mathbf{C}_1 |X)=\prod_i p_{\theta_{LLM}}(c_1^i |X, \mathbf{C}_1^{<i})$ autoregressively. The objective has thus become: 
\begin{equation}
    \mathcal{L}_{\mathit{LLM}}=-\sum_{i=1}^{T'} \log p_{\theta_{\mathit{LLM}}}(c_1^i |X,\mathbf{C}_1^{<i}),
    \label{eq:llm:loss}
\end{equation}  
where $T'$ is the number of acoustic tokens generated by LLM, $\theta_{\mathit{LLM}}$ is the parameter of LLM, $c_{1}^i$ is the token generated at step $i$, $\mathbf{C}_1^{<i}$ are previous tokens, and $X$ is the text or video condition. 

During inference, the LLM will autoregressively predict the next token until \textit{<eos>} is generated. Our LLM thus serves as the bridge for connecting between different pairs of modalities. The generated output will be decoded subsequently. 

Due to the limited computation resource available, we use Gemma-2B~\cite{gemma}, a lightweight open-source LLM developed by Google, which is claimed to have comparable performance with LLaMA-2-7B~\cite{llama2} on many QA and reasoning tasks. We use Low-Rank Adaptor (LoRA)~\cite{lora} to finetune Gemma to make it understand vision/text conditions and generate audio tokens. 
 
\subsection{Non-Autoregressive Transformer for Audio Refinement}
In C3LLM, we propose an audio refinement method to further ensure the generated audio fidelity. Inspired by~\cite{valle}, we utilize a non-autoregressive Transformer (NAR) to transform coarse acoustic tokens from LLM's output to fine-grained acoustic tokens. We treat the video embedding or text input as condition and concatenate it with the generated coarse acoustic tokens. In the original paper, the NAR is used to predict seven layers of acoustic tokens given by the first layer. However, we find this design very slow to converge during our experiment. We adopt a simpler design to only utilize two layers of codebooks, and train the NAR to predict the second layer given the first layer prediction generated by LLM. Thus probability distribution for the next layer is given by $p_{\theta_{\mathit{NAR}}}(\mathbf{C}_{2}|\mathbf{X}, \mathbf{C}_{1} )$, and we want to minimize the objective function:
\begin{equation}
\mathcal{L}_{\mathit{NAR}} = -\log p_{ \theta_{\mathit{NAR}}}(\mathbf{C}_{2}|\mathbf{X}, \mathbf{C}_{1})= -\sum_{i=1}^{T'} \log p_{\theta_{\mathit{NAR}}}(c_{2}^i | \mathbf{X}, \mathbf{C}_{1} ).
\end{equation}

\subsection{Detokenization for High Fidelity Output}
The decoder combines LSTM and CNN architectures. The LSTM component emphasizes temporal consistency, while the CNN component reconstructs frequency information. The model employs a combination of L1 loss for the time domain and a set of L1 and L2 losses for the Mel-spectrogram in the frequency domain, across various time scales. To effectively preserve audio information, the model incorporates two strategies: 
1) Utilizing a greater number of tokens to represent each second of audio, thereby increasing the sample rate.
2) Employing multiple codebooks to capture a wider bandwidth, enhancing the representation.

The audio waveform can experience interference from multiple sources. By employing more codebooks, the model can effectively decompose overlapping signals, with each codebook capturing audio of different frequencies.

\section{Experiments}

\subsection{Training Datasets}
We apply corresponding datasets for the three tasks. For video-to-audio task, we finetune our model on the VGGSound~\cite{vggsound} dataset, which contains over 310 classes of 200,000+ videos, capturing challenging real-world acoustic scenarios. Concerning the large size, we use around half of the common version of VGGSound containing 164 classes. The resulting number of training video samples is 63,853.

For audio-to-text and text-to-audio tasks, we use the AudioCaps~\cite{audiocaps} dataset. AudioCaps~\cite{audiocaps} dataset is a large-scale dataset of about 46K audio clips paired with human-written text collected via crowdsourcing on the AudioSet~\cite{Audioset} dataset. The dataset contains a diverse range of 10-second audio samples from various natural sources, including vehicles, animals, weather, etc. We filter all the failed links and produce 45,028 sound files in train split. For audio-to-test task, we notice that in some papers such as EnCLAP~\cite{enclap}, multiple referencing ground truth are used for evaluation. As the original AudioCaps paper~\cite{audiocaps} only contains one caption per audio clip, we choose to only use one caption as the referencing ground truth.

\subsection{Evaluation Metrics}
The evaluation metrics are summarized as follows: For video-to-audio and text-to-audio tasks, we use the Inception score (ISc) and Frechet audio distance (FAD)  to evaluate audio fidelity. For audio-video relevance, we utilize the MKL metric~\cite{specvqgan} and we use KL for text-to-audio task. For audio-to-text task, we use the CIDEr (Consensus-based Image Description Evaluation), SPIDEr (SPeech-to-Image Description Evaluation), and SPICE (Semantic Propositional Image Caption Evaluation).
Furthermore, to evaluate the synchronization of the generated audio in the video-to-audio setting, we use the same evaluation metrics as CondFoleyGen~\cite{Conditional-floey}, namely \# Onset Accuracy~\cite{Conditional-floey}, and Onset AP~\cite{Conditional-floey}.

\subsection{Evaluation and comparison}
We mainly compare our model with CoDi~\cite{codi}, which is the current state-of-the-art model combining different multimodal content generation tasks. We download the pretrained fp16 version of CoDi~\cite{codi} model and evaluate on the same test set. The training and evaluation are conducted on NVIDIA GeForce RTX 4090. The main result is presented in Table ~\ref{tab:main}

\begin{figure}
    \centering
    \includegraphics[width=\linewidth]{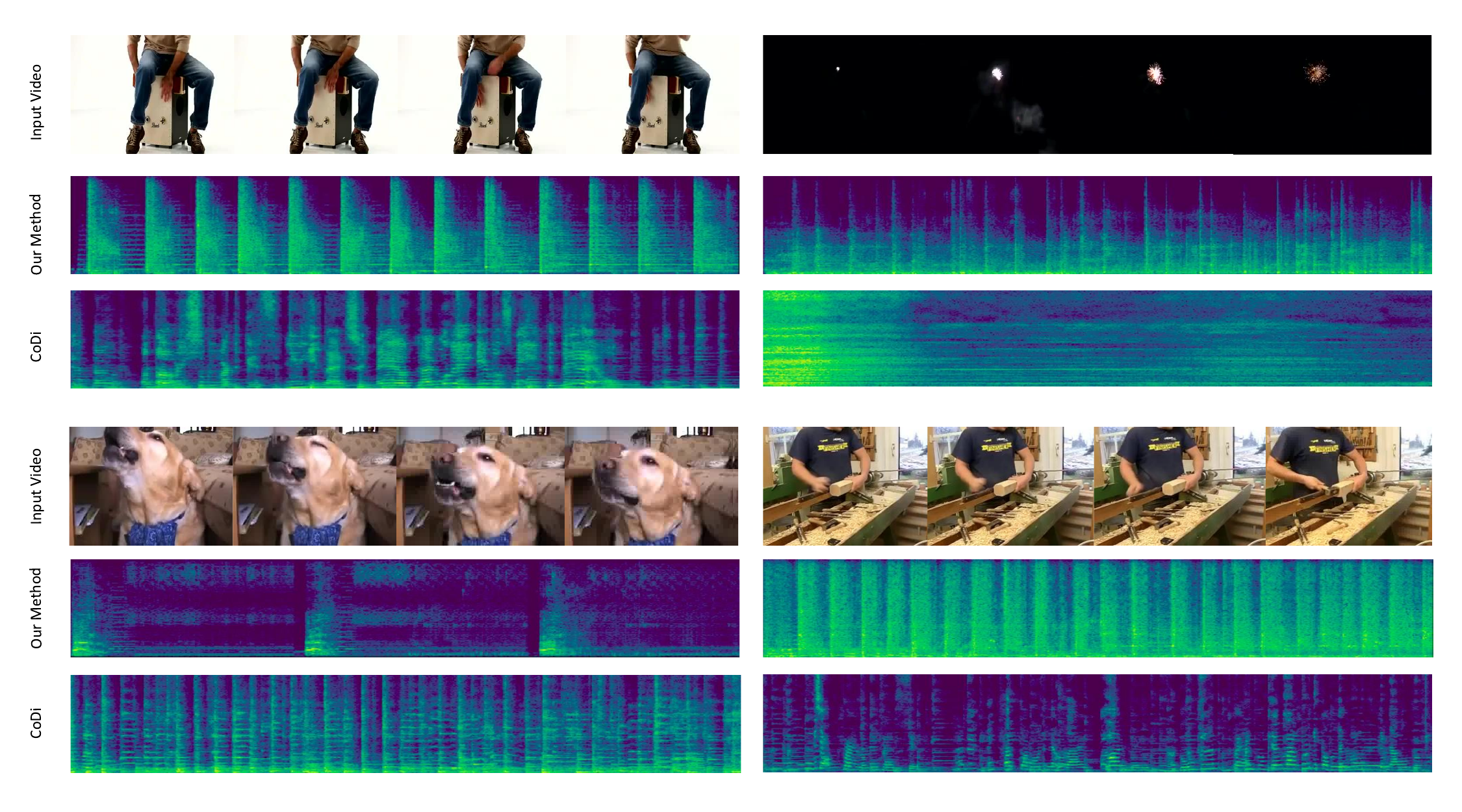}
    \vspace{-0.1in}
    \caption{ Comparison with baseline for video-to-audio generation task. CoDi failed to generate semantic-aligned audio and the generation is not clean, often mixed with human speaking or noise. Our method can produce aligned audio with clear synchronization.
    }
    \label{fig3:sync}
    \vspace{-0.1in}
\end{figure}

\begin{table*}[t]
    \centering
    \setlength\tabcolsep{5pt} 
    \begin{tabular}{ccccc}
        Task & Method & \multicolumn{3}{c}{Metric}  \\
        \toprule
        \multirow{4}{*}{V2A} 
        &  & \cellcolor[rgb]{1, 1, 0.9}{KL}$\downarrow$  & \cellcolor[rgb]{1, 1, 0.9}{ISc}$\uparrow$  & \cellcolor[rgb]{1, 1, 0.9}{FAD}$\downarrow$  \\
        \cline{3-5}
        & Codi~\cite{codi} & 3.6449  & \textbf{3.6941}  & 11.5036\\
        & Ours (w/o NAR) & 3.5873  & 2.1301  & 17.1095\\
        & Ours & \textbf{3.5522} & 2.5783 & \textbf{10.2217} \\
        \hline
        \multirow{3}{*}{A2T} 
        &  & \cellcolor[rgb]{1, 1, 0.9}{SPIDEr}$\uparrow$ & \cellcolor[rgb]{1, 1, 0.9}{CIDEr}$\uparrow$ & \cellcolor[rgb]{1, 1, 0.9}{SPICE}$\uparrow$  \\
        \cline{3-5}
        & Codi~\cite{codi} & 0.0726 & 0.0812 & 0.0640 \\
        & Ours &  \textbf{0.3150} & \textbf{0.4721} & \textbf{0.1579}\\
        \hline
        \multirow{4}{*}{T2A} 
        &  & \cellcolor[rgb]{1, 1, 0.9}{KL}$\downarrow$  & \cellcolor[rgb]{1, 1, 0.9}{ISc}$\uparrow$  &  \cellcolor[rgb]{1, 1, 0.9}{FAD}$\downarrow$  \\
        \cline{3-5}
        & CoDi~\cite{codi} & \textbf{3.1786}  & \textbf{4.5047} & \textbf{10.1597} \\
        &  Ours (w/o NAR) & 3.4865 & 2.5732  & 22.7320 \\& Ours  & 3.6765  & 2.9606  & 18.1615\\

        \bottomrule
    \end{tabular}
    \caption{Quantitative comparison with baseline on three tasks.}
    \label{tab:main}
\end{table*}

\begin{table*}[t]
    \centering
    \setlength\tabcolsep{5pt} 
    \begin{tabular}{cccc}
    Method & \multicolumn{2}{c}{Metric}  \\
    \toprule
    & \cellcolor[rgb]{1, 1, 0.9}{\# Onset Accuracy}$\uparrow$  & \cellcolor[rgb]{1, 1, 0.9}{Onset AP}$\uparrow$    \\
    \cline{2-3}
    Codi~\cite{codi} & 0.097&  0.535\\
    Ours & \textbf{0.142} & \textbf{0.670}  \\
    \bottomrule
    \end{tabular}
    \caption{Results for evaluating video-audio synchronization on VGGSound dataset}
    \label{tab:synchronization}
\end{table*}

\begin{table*}[htbp]
    \centering
    \small
    \setlength\tabcolsep{2pt}
    \begin{tabular}{ | c | c | c | }
        \hline
        Ground truth & Our method & CoDi~\cite{codi} \\ \hline
        continuous snoring of a person
        &
        a person snoring
        & 
        dog sleeping on a bed
        \\ 
        \hline
        church bells ringing
        &
        bells are ringing
        & 
        angel is in a coat by in red hat
        \\ 
        \hline
        a car engine is revving while driving
        &
        a vehicle engine accelerates
        & 
        driving for speed going for a crossing
        \\ 
        \hline
        a telephone ringing
        &
        a telephone rings several times
        & 
        phone of the phone
        \\ 
        \hline
        
        A cat meowing a few times
        &
        a cat is meowing
        & 
        \shortstack{catwoman's cats are coming cat of \\
        the cats in a cat calendar}
        \\ 
        \hline
        spinning tires on pavement
        &
        a car is skidding wildly
        & 
        \shortstack{car for just avoiding a speeding car \\ wreck with cold scary highway blare}
        \\ 
        \hline
  \end{tabular}
  \caption{Samples output obtained from AudioCap test set. We observe that CoDi often fails to capture the semantic meaning, and the generated captions are more like describing visual input rather than acoustic sound. Additionally, the output is not sufficiently fluent. We thus believe LLM structure is more capable for captioning tasks }\label{tbl:captioning-result}
\end{table*}

\begin{table}[!th]
\centering%
\begin{minipage}[b]{0.48\textwidth}
\centering%
\caption{Additional A2T task conducted on Clotho dataset}\label{tab:clotho}
    \begin{tabular}{cccc}
    Method & \multicolumn{3}{c}{Metric}  \\
    \toprule
    & \cellcolor[rgb]{1, 1, 0.9}{SPIDEr}$\uparrow$ & \cellcolor[rgb]{1, 1, 0.9}{CIDEr}$\uparrow$ & \cellcolor[rgb]{1, 1, 0.9}{SPICE}$\uparrow$  \\
    \cline{2-4}
    Codi~\cite{codi} & 0.0640 & 0.0766 & 0.0514 \\
    Ours &  \textbf{0.2088} & \textbf{0.3097} & \textbf{0.1078}\\
    \bottomrule
    \end{tabular}
\end{minipage}%
\hspace{3mm}%
\begin{minipage}[b]{0.48\textwidth}
\centering%
\caption{Additional V2A task conducted on VAS dataset}\label{tab:vas}
    \begin{tabular}{cccc}
    Method & \multicolumn{3}{c}{Metric}  \\
    \toprule
    & \cellcolor[rgb]{1, 1, 0.9}{KL}$\downarrow$  & \cellcolor[rgb]{1, 1, 0.9}{ISc}$\uparrow$  & \cellcolor[rgb]{1, 1, 0.9}{FAD}$\downarrow$   \\
    \cline{2-4}
    Codi~\cite{codi} & 4.54874&  \textbf{3.12170} &  11.80060 \\
    Ours & \textbf{3.97517} &  2.69774&  \textbf{10.35411}  \\
    \bottomrule
    \end{tabular}
\end{minipage}
\end{table}


For audio-to-test task, we use the open-sourced Audio Captioning metrics~\cite{aac_metric} for evaluation. we observe that CoDi's performance is lower by a large margin, which might be the result that LLM is more capable of captioning tasks due to the next-word prediction method. We include some test results in Table~\ref{tbl:captioning-result}. 

For video-to-audio task, our result is better than the baseline except for the ISc metric. The ISc metric measures the diversity of the generated audio. As our model is only fine-tuned on Gemma-2B~\cite{gemma} backbone, we believe our result will be improved with more power backbone and tunable parameters, given more training resources. Additionally, our method achieves synchronization with the input video, as shown in Figure~\ref{fig3:sync}. The quantitative evaluation is presented in Table ~\ref{tab:synchronization}

We notice that for text-to-audio, CoDi is trained on multiple datasets such as AudioCaps~\cite{audiocaps}, AudioSet~\cite{Audioset}, BBC Sound Effect, Soundnet~\cite{soundnet}, and Freesound. On the other hand, our model is only trained on AudioCaps. The total number of training examples for CoDi is significantly larger than ours. Furthermore, there is a huge domain gap between text and audio. Audio waveform has time information while text does not, so it is hard to map a sentence to a specific audio token in each time frame. Besides, we utilize LoRA~\cite{lora} to finuetune LLM which is not capable of bridging the gap with so few trainable parameters. These are the reasons for the comparison result.

\subsection{Ablation Studies}
We hereby conduct experiments to test how the non-autoregressive transformer will refine the output coarse acoustic tokens. We include our results in Table~\ref{tab:main}. As shown in the table, the NAR plays a central role in further improving the result. 

To mitigate the possible effect of using only one referencing ground truth for evaluating audio-to-text task, we tested our model on the Clotho dataset~\cite{clotho}. The Clotho dataset contains 4981 audio clips and 3938 clips in train split. Audio waveforms are from 15 to 30s duration, and each audio has 5 captions which are 8 to 20 words long. We perform similar processing as the AuidoCaps dataset for training and evaluation. Table~\ref{tab:clotho} tabulates the results.

We also conducted an additional evaluation of our model on the VAS~\cite{vas} dataset for video-to-audio task, as shown in Table ~\ref{tab:vas}. We obtain a similar result as VGGSound. Our model continues to outperform the baseline.

\section{Conclusion and Discussion}
\subsection{Limitation and Future Work}
Due to limited computation power, we can only take one modality input and generate text or audio. Next, we want to condition on two or more modalities and generate video, and we want to generate long audio/video. Moreover, to bridge the modality gap between text-to-audio generation, we will adopt semantic tokens of audio to give well-aligned information. We will utilize more powerful LLM backbones such as LLama3~\cite{llama3}

\subsection{Conclusion}
In this paper, we present C3LLM, a unified structure that can perform three tasks namely video-to-audio, audio-to-text and text-to-audio. Our model capitalizes on the power of LLM for translating and aligning between different modalities. We also propose a non-autoregressive transformer for audio refinement. Through extensive experiments, we show that our model can synthesize high-fidelity audio, ensuring semantic alignment with input, especially synchronization with the visual condition. 

Although our model demonstrates excellent results through evaluation, challenges exist that can restrain the performance. Specifically for the audio-to-text task, our model relies on the pre-trained CLAP encoder that poses an upper-bound for modeling more complex scenarios. A more efficient way for audio encoding is worthwhile in future research.

\bibliographystyle{plain}
\bibliography{Styles/reference}

\newpage
\appendix

\section{Appendix / supplemental material}
\subsection{Audio-to-Text examples}
In this section, we provide more examples of the predicted output of the audio-to-text task in Table~\ref{tbl:additional-captioning-result}, in addition to Table \ref{tbl:captioning-result}. We also include some examples from the Clotho dataset~\cite{clotho}, shown in Table~\ref{tbl:additional-captioning-result-clotho}.

\begin{table*}[htbp]
    \centering
    \small
    \setlength\tabcolsep{2pt}
    \begin{tabular}{ | c | c | c | }
        \hline
        Ground truth & Our method & CoDi~\cite{codi} \\ \hline
        a person snoring
        &
        a person snores lightly
        & 
        sleeping or a baby in a skiing
        \\ 
        \hline
        a river stream of water flowing
        &
        water is rushing by
        & 
        brown birds outside in the forest
        \\ 
        \hline
        clicking followed by humming noise
        &
        an engine humming and clicking
        & 
        highway worker hiking cold turkey
        \\ 
        \hline
        \shortstack{sounds of a river with man \\
        briefly mumbling}
        &
        water is flowing and a man speaks
        & 
        waterfall falls in water
        \\ 
        \hline
        
        several goats bleat
        &
        sheep bleat nearby
        & 
        the brown goat
        \\ 
        \hline
        a police siren going off
        &
        a siren is emitted
        & 
        \shortstack{clear weather warning direction as \\water officer does traffic safety \\
        direction while a traffic stop in\\
        her direction.}
        \\ 
        \hline
  \end{tabular}
  \caption{Additional samples output obtained from AudioCap test set. We again notice that words describing colors or visual scenes exist in CoDi's output which is unusual.}\label{tbl:additional-captioning-result}
\end{table*}

\begin{table*}[htbp]
    \centering
    \small
    \setlength\tabcolsep{2pt}
    \begin{tabular}{ | c | c | c | }
        \hline
        Ground truth 1 & Our method & CoDi~\cite{codi} \\ \hline
        \shortstack{a radio dispatcher and an \\
        officer are communicating \\
        over the radio}
        &
        \shortstack{a radio is tuned, as a \\
        person speaks over the radio}
        & 
       \shortstack{foreign radio is not time to \\
       watch someone on the phone to\\
       communications wire}
        \\ 
        \hline
        \shortstack{lost of people are conversing in \\
        a very busy diner}
        &
        \shortstack{a group of people are talking \\
        and laughing}
        & 
        people walking in the area
        \\ 
        \hline
        \shortstack{a machine is running in a \\
        humming manner \\
        while metal is buzzing}
        &
        \shortstack{a buzzing electric engine \\
        that is trying to start up}
        & 
        \shortstack{pilot cable bowler take a caution \\steer light not flying < speed> engine.}
        \\ 
        \hline
  \end{tabular}
  \caption{Additional samples output obtained from Clotho test set. Due to the space limit, we only include ground truth caption 1. Other referencing groundtruth can be found in the CSV file provided by the original paper~\cite{clotho}.}\label{tbl:additional-captioning-result-clotho}
\end{table*}

\end{document}